\documentclass{article}

\usepackage{microtype}
\usepackage{graphicx}
\usepackage{subfigure}
\usepackage{booktabs} 
\usepackage{array}
\usepackage{amsmath}

\usepackage{hyperref}

\usepackage[accepted]{icml2019}

\icmltitlerunning{Virtual Conditional Generative Adversarial Networks}

\begin{document}

\twocolumn[
\icmltitle{Virtual Conditional Generative Adversarial Networks}



\icmlsetsymbol{equal}{*}

\begin{icmlauthorlist}
\icmlauthor{Haifeng Shi}{}
\icmlauthor{Guanyu Cai}{}
\icmlauthor{Yuqin Wang}{}
\icmlauthor{Shaohua Shang}{}
\icmlauthor{Lianghua He}{}
\end{icmlauthorlist}



\icmlkeywords{Machine Learning, ICML}

\vskip 0.3in
]




\begin{abstract}
When trained on multimodal image datasets, normal Generative Adversarial Networks (GANs) are usually outperformed by class-conditional GANs and ensemble GANs, but conditional GANs is restricted to labeled datasets and ensemble GANs lack efficiency. We propose a novel GAN variant called virtual conditional GAN (vcGAN) which is not only an ensemble GAN with multiple generative paths while adding almost zero network parameters, but also a conditional GAN that can be trained on unlabeled datasets without explicit clustering steps or objectives other than the adversary loss. Inside the vcGAN’s generator, a learnable ``analog-to-digital converter (ADC)" module maps a slice of the inputted multivariate Gaussian noise to discrete/digital noise (virtual label), according to which a selector selects the corresponding generative path to produce the sample. All the generative paths share the same decoder network while in each path the decoder network is fed with a concatenation of a different pre-computed amplified one-hot vector and the inputted Gaussian noise.  We conducted a lot of experiments on several balanced/imbalanced image datasets to demonstrate that vcGAN converges faster and achieves improved Frechét Inception Distance (FID). In addition, we show the training byproduct that the ADC in vcGAN learned the categorical probability of each mode and that each generative path generates samples of specific mode, which enables class-conditional sampling. Codes are available at \url{https://github.com/annonnymmouss/vcgan}
\end{abstract}

\section{Introduction}
\label{introduction}

Generative Adversarial Network(GAN)~\cite{goodfellow2014generative} is a generative model composed of two neural networks which are trained in opposite directions. Since it was proposed in 2014, GAN has quickly become one of the research hotspots in the field of  deep learning and artificial intelligence, and has been widely used in tasks including image generation, image style transfer and representation learning.

The original GAN has the problems of unstable training and mode collapse~\cite{goodfellow2016nips}, which affects the fidelity of the generator distribution. To improve the generator distribution, many methods have been proposed and they can be roughly divided into three categories: (1) Improving the training approach, such as Unrolled GAN~\cite{metz2016unrolled} or WGAN-GP~\cite{gulrajani2017improved}, etc., to overcome the problem of mode dropping by stabilizing GAN's training; (2) Using label conditioning, such as the conditional LAPGAN~\cite{denton2015deep}, AC-GAN~\cite{odena2017conditional}, and cGAN~\cite{mirza2014conditional}, which can always significantly improve the sample quality.~\cite{goodfellow2016nips,salimans2016improved}; (3) Ensembling multiple GANs, such as MGAN~\cite{hoang2017multi}, AdaGAN~\cite{tolstikhin2017adagan}, Mix+GAN~\cite{arora2017generalization}, and MAD-GAN~\cite{ghosh2018multi}, to cover more modes and improve the fidelity of generator distribution. However, conditional GANs can not be trained on unlabeled data sets. The ensemble GANs usually significantly increases the number of parameters and training time, which is cumbersome.

This paper proposed a GAN called Virtual Conditional GAN(vcGAN) --- It is both a conditional GAN which supports class-conditional sampling and a type of ensemble GAN that contains multiple generative paths(sub-generators). Unlike other conditional GANs trained only on labeled datasets and ensemble GANs which increase model size a lot, we elaborately designed vcGAN's generator architecture so that it can be trained on unlabeled datasets and adds zero or negligible trainable parameters to single-generator GANs but inherits the merits of conditional GANs and ensemble GANs. The vcGAN has an Analog-to-Digital Convertor (ADC) module converting part of the inputted Gaussian noise to digital noise(virtual label), which injects step signal and provides discontinuity to the generator. As a result, the function space of the generator is enlarged so that multimodal data distributions can be better modeled by vcGAN than previous GANs with continuous generators. Moreover, the ADC supports to learn to yield virtual labels of proper categorical probabilities, ensuring vcGAN's performance on both class-balanced and imbalanced datasets. The proposed method can be integrated into any GAN variants easily.

\section{Related Work}
\label{related work}
Algorithms of the conditional GAN family trained with label information can usually generate samples of higher quality than those of normal unconditional GANs.~\cite{goodfellow2016nips,salimans2016improved}. To our knowledge, conditional GAN (cGAN)~\cite{mirza2014conditional} is the first GAN model which utilizes label information. Input of cGAN's generator is the combination of a noise vector and a label vector, whereas input of its discriminator is a tuple consists of a real or generated sample and the corresponding label. Discriminator gives the probability that the input sample is drawn from real data distribution. In AC-GAN~\cite{odena2017conditional}, in addition to judging whether the input sample is true or false, discriminator should also predict the category the sample belongs to. The goal of generator is to generate samples that discriminator considers to be real and reduce the classification error of discriminator under given noises and categories. Experiments on ImageNet~\cite{russakovsky2015imagenet} demonstrate that AC-GAN generates samples with higher quality and diversity than original GAN. Conditional LAPGAN~\cite{denton2015deep} also exhibits that samples generated with label information appear more object-like and have more clearly defined edges.

However, conditional GANs are restricted to labeled dataset, which limits its application. In order to break through the limit, some GAN variants that synthesize category labels of training samples in an unsupervised manner are proposed, such as CatGAN~\cite{springenberg2015unsupervised}, InfoGAN~\cite{chen2016infogan} and SplittingGAN~\cite{grinblat2017class}. CatGAN focuses on improving the classification accuracy of discriminator rather than the generation aspect. InfoGAN learns disentangled representations of datasets in an unsupervised manner by maximizing the mutual information between generated samples and the latent code. SplittingGAN utilizes k-means algorithm to cluster the representation of training-set samples in the last hidden layer of discriminator network and labels these training-set samples for further AC-GAN training, which improves the quality of generated images remarkably.

Another way to improve GAN's generation, especially to overcome the mode dropping problem, is to integrate multiple generators. Standard Ensemble of GANs~\cite{wang2016ensembles} simply trains N different randomly initialized GAN models and randomly selects a generator to generate samples, while Cascade of GANs~\cite{wang2016ensembles} and AdaGAN~\cite{tolstikhin2017adagan} trains generators in sequence and gradually adding incremental components to the current model distribution. It is worth mentioning that generators of the above three GAN models are not trained simultaneously, which may cover more modes as well as more poor samples step by step. Ensemble GAN models that train generators simultaneously include Mix+GAN~\cite{arora2017generalization}, MAD-GAN~\cite{ghosh2018multi}, MGAN~\cite{hoang2017multi}, MEGAN~\cite{park2018megan}, and DeLiGAN~\cite{gurumurthy2017deligan}. Mix+GAN has multiple generators and discriminators that have independent neural network parameters and learnable mixed weights, which is computational expensive. MGAN reduces the model size to some degree by parameter sharing. MGAN also adds extra classification loss term in the generator's loss function to force each generator to specialize on different modes in the training set.~\cite{hoang2017multi} pointed out that the mixture weights of the generator distributions of MGAN are unreasonably fixed and evenly distributed, and proposed Mixture of Experts GAN (MEGAN). Inside the MEGAN, a learnable Gating Networks based on Straight-Through Gumbel Softmax~\cite{jang2016categorical} picks one sample from all the generated samples of multiple generators as the output of the model. MEGAN's Inception Score is comparable to MGAN on the CIFAR-10 dataset and is superior to other single-generator GAN models. To the best of our knowledge, DeLiGAN is currently the most lightweight GAN which can be regard as an ensemble GAN. It replace common generator's input with trainable mixture of Gaussians to improve the model's preformance when the dataset is small but diverse.

%
%
%
%

\section{Virtual Conditional GAN}
\label{Virtual Conditional GAN}


Previous works add label information to train conditional GANs and ensemble multiple GANs to improve sample quality and to alleviate model collapse problem of ordinary GANs. We find the common consequence of the two modifications is that they enable the generative model to learn multimodal distribution supported on several disconnected sets, which is intractable for ordinary GANs with a continous generator function and a input of Gaussian noise. The label inputted to conditional GAN's family can be regarded as digital/discrete noise which provide impulse or step signal to the generator to cut off the model's manifold. Model of ensemble GAN's family contains the structure of  multiple \emph{generative paths} and a selector to form piecewise continous function that is suitable for multimodal datasets. Inspired by above, we design a novel generator architecture combined with discrete label generation and multiple generative paths, and propose virtual conditional generative adversarial network (vcGAN).
\subsection{Generator's Architecture}



\begin{figure}[ht]
\begin{center}
\centerline{\includegraphics[width=\columnwidth]{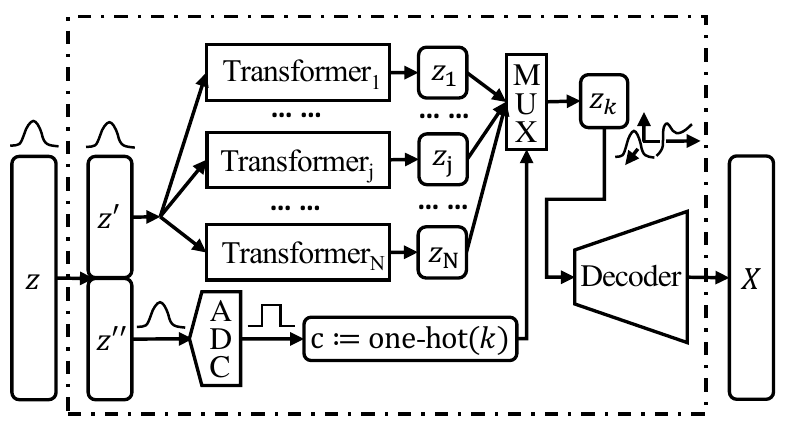}}
\caption{The generator's architecture of vcGAN.}
\label{generator}
\end{center}
\vskip -0.2in
\end{figure}

The high-level architecture of vcGAN is the same as standard GANs, without labels input or auxiliary classifier added. However, vcGAN's generator is elaborately designed as Figure~\ref{generator} illustrates and is not a continuously differentiable neural network for the input noise. The generator is composed of  a learnable analog-to-digital converter (ADC), a selector (multiplexer or MUX), a decoder network/renderer $R(\cdot)$ and $N$ noise transformers. Every noise transformer along with the MUX and the shared decoder network $R(\cdot)$ forms a \emph{generative path} or sub-generator so that the entire generator can be viewed as  an ensemble of sub-generators.

The input is a $M$-d multivariate Gaussian noise vector which is split into two slices, $z'$ and $z''$ inside the generator. The dimension of $z'$ and $z''$ is $L=M-N$ and $N$, respectively. $z'$ is converted into $N$ different random vectors $z_1,z_2,...,z_j,...,z_N$ by $N$ noise transformers $T_1,T_2,...,T_j,...,T_N$, i.e. $z_j=T_j(z')$

ADC converts $z''$ (which can be regarded as analog noise) into a one-hot vector $c$ (which can be regarded as digital noise or virtual label ) of $N$ categories. In this work the ADC is implemented by utilizing the Gumbel-Max Trick\cite{jang2016categorical,maddison2014sampling} and the \emph{inverse transform sampling} method nestedly:
\begin{align}
  & c=\text{one-hot}(k)=\underbrace{[0,...,0,}_{k-1}1,\underbrace{0,...,0}_{N-k}{{]}^{\text{T}}} \\ 
 & k=\underset{i\in \{1,...,N\}}{\mathop{\arg \max }}\,\left( \log \left( {{p}_{i}} \right)-\log \left( -\log \left( {{\Phi }^{\text{-1}}}\left( {{{{z}''}}_{i}} \right) \right) \right) \right)  
\end{align}
where $p_i$ is the probability that ADC outputs one-hot($i$), and can be predefined or learnable. ${{{{z}''}}_{i}}$ is the $i$th element of ${{z}''}$. $\Phi^\text{-1}(\cdot)$ is the inverse of the standard normal CDF(cumulative density function. $c$ controls the MUX to select $z_k$ for the renderer $R(\cdot) $from $z_1,z_2,...,z_j,...,z_N$. The selection operation of MUX can be expressed by matrix multiplication as:
\begin{align}
{{[{{z}_{k}}]}_{L\times 1}}={{[{{z}_{1}},{{z}_{2}},...,{{z}_{j}},...,{{z}_{N}}]}_{L\times N}}\cdot {{[c]}_{N\times 1}} 
\end{align}
$R$ is a continuously differentiable mapping that maps $z_k$ to the final generated sample $x$, i.e. $x=R(z_k)$

In general, noise transformers and decoder network $R$ in  Figure~\ref{generator} can be neural networks or other functions to form variants of DeLiGAN or MGAN. Particularly, in vcGAN's generator, $R$ is the only neural network (implemented by DCGAN-like generator ~\cite{radford2015unsupervised}), and transformers are not trainable, to extremly reduce the computational cost. The $j$th noise transformer $T_j$ concatenates a scaled and biased vector $c_j$ to $z'$, outputing $z_j$ as follows:
\begin{align}
&z_j=\left[
\begin{matrix}
z'\\
c_j\\
\end{matrix}
\right]
=\left[
\begin{matrix}
z'\\
\underbrace{[b,...,b,}_{j-1}A,\underbrace{b,...,b}_{N-j}{{]}^{\text{T}}}\\
\end{matrix}
\right]\nonumber\\
& {{c}_{j}}=b+(A-b)\cdot \text{one-hot}(j)
\end{align}
where $A$ and $b$ are constant value, and the detailed explaination is given in Section~\ref{scalebias}. The design of noise transformers in this way eliminates the computation of MUX, and simplifies the mapping of vcGAN generator as:
\begin{align}
  & G(z)=G({z}',{z}'')=R({{z}_{k}})=R\left( \left[ \begin{matrix}
{{z}'} \\ 
{{c}_{k}} \\ 
\end{matrix} \right] \right)\nonumber\\
&=R\left( \left[ \begin{matrix}
{{z}'} \\ 
b+(A-b)\cdot \text{one-hot}(k) \\ 
\end{matrix} \right] \right)
\end{align}

It is obvious that if $p_i$ is a constant (for example, $\frac{1}{N}$), there are no increased trainable parameters for vcGAN compared with normal GANs. Enabling $p_i$s to learn only adds $N$ inevitable parameters. Hence we claim that vcGAN is the most economical one in the family of ensemble GANs.

Noise transformers, MUX and $R$ are continuous functions of their inputs, while the ADC has a discontinuity. Therefore, the vcGAN generator is not a continuous function and has some point of infinite gradients, which is the great difference between vcGAN and other traditional GAN models that require generators to be continuously differentiable. A continuously differentiable generator can only convert the input multivariate Gaussian distribution into a distribution whose support set is a connected set, however, the distribution of a real training datasets is not necessarily connected, leading to the result that a continuous generator can not perfectly learn such a distribution, which affects the quality of the generated samples. 

In theory, vcGAN can not only learn such kind of distributions but also performs better because of its lager function space. When the training set data distribution is a connected set, vcGAN can also degenerate to a continuous and differentiable traditional generator, you only need to reduce the weight of the $c_k$ in $R$.

\subsection{Scale and Bias of the One-hot Vector}
\label{scalebias}

If the splicing of the Gaussian noise and the one-hot vector is input into $R$ without any processing, $R$ is likely to ignore the one-hot part, which we think is because the amplitude of one-hot vector is too small compared to that of Gaussian noise, leading to the fact that the semantic information of original one-hot vector is not significant.

Intuitively, we want to scale and bias the one-hot vector so that any two composite noise vectors from different class $i$, $j$ satisfy the condition that the maximum value of the inter-class distance is not less than the minimum value of the intra-class distance:
\begin{align}
  & \min ({{d}_{\text{inter}}})\ge \max ({{d}_{\text{intra}}}) \label{eq7}\\
 & {{d}_{\text{intra}}}=Euclidean~(z_{i}^{(1)},z_{i}^{(2)}) \nonumber\\
 & {{d}_{\text{inter}}}=Euclidean~({{z}_{i}},{{z}_{j}}) \nonumber
\end{align}
Due to unboundedness of Gaussian noise, we take the maximum and minimum by $\pm\delta$ times the standard deviation, and (\ref{eq7}) is proxyed by:
\begin{align}
E[{{d}_{\text{inter}}}]-\delta \cdot std[{{d}_{\text{inter}}}]=E[{{d}_{\text{intra}}}]+\delta \cdot std[{{d}_{\text{intra}}}]\label{eq8} 
\end{align}
Meanwhile, we prefer a zero-sum amplified one-hot vector and normalize it by (\ref{eq9}):
\begin{align}
(N-1)b+A=0\label{eq9}
\end{align}
Combining (\ref{eq8}) and (\ref{eq9}), a very accurate approximate solution of the equations is given (see Appendix for details):
\begin{align}
&\left\{ \begin{matrix}
A=(1-N)\cdot b \\ 
b=-h/N \\ 
\end{matrix} \right. \label{eq10} \\
 \text{where }h=&\sqrt{\frac{1}{8}{{\left( \sqrt{{{v}^{2}}+\delta \sqrt{32L}}+v \right)}^{2}}-L} \nonumber\\ 
 v=&\sqrt{2L}+\delta \sqrt{1-\frac{1}{8L}}\text{ }-\text{ }\frac{1}{\sqrt{8L}} \nonumber
\end{align}

(\ref{eq10}) is the formula for the noise transformers in vcGAN to scale and bias the one-hot vectors. Generally, setting the hyper-parameter $\delta$ to 2 or 3 is probably enough to satisfy (\ref{eq7}). The experiments in Section~\ref{experiments} demonstrate the robustness of hyper-parameter values.

\subsection{Learn Categorical Probabilities of the ADC}

As with previous GANs, vcGAN adopts gradient descent algorithm to train and support original GAN objective function, wgan-gp objective function or other types of objective function. If there is no need to train $p_i$ for each category (for example, we have the prior that all categories are uniformly distributed), $p_i$s are constant values and only generator and discriminator are trained. In this setting, our model is called vcGAN-FP (fixed $p$).

However, if the proportion of all categories of samples in the training set is unknown or unbalanced, our model can also learn $p_i$s of all categories. In this setting, our model is called vcGAN-LP (learnable $p$). $p_i$s are parameterized by $N$ trainable variables $q_1,q_2,...,q_N$ belonging to ADC and converted to multi-class probability distribution through softmax function.


The objective/loss function of the generator is as follows (let's take Wasserstein GAN's loss for example):
\begin{align}
&los{{s}_{G}}(q,{{\theta }_{R}};D)={{E}_{z\sim{\ }{{p}_{z}}}}\left[ -D(G(z)) \right] \nonumber\\ 
 &={{E}_{{z}''\sim{\ }{{p}_{{{z}''}}}}}\left[ {{E}_{{z}'\sim{\ }{{p}_{{{z}'}}}}}\left[ -D\left( G\left( {z}',{z}'' \right) \right) \right] \right] \nonumber\\ 
 &=\sum\limits_{j=1}^{N}{\left[ {{p}_{j}}\cdot {{E}_{{z}'\sim{\ }{{p}_{{{z}'}}}}}\left[ -D\left( R\left( \left[ \begin{matrix}
{{z}'} \\ 
{{c}_{j}} \\ 
\end{matrix} \right] \right) \right) \right] \right]} \nonumber\\ 
  &\simeq \sum\limits_{j=1}^{N}{\left[ \frac{{{-e}^{{{q}_{j}}}}}{\sum\limits_{h=1}^{N}{{{e}^{{{q}_{h}}}}}}\cdot \frac{1}{{{B}_{j}}}\sum\limits_{i=1}^{{{B}_{j}}}{\left[ D\left( R\left( \left[ \begin{matrix}
{{{{z}'}}^{(i)}} \\ 
{{c}_{j}} \\ 
\end{matrix} \right] \right) \right) \right]} \right]}
\end{align}
where $\theta_R$ denotes parameters of $R$, $B_j$ is the number (unfixed) of samples that belong to class $j$ in a batch, and batchsize $=\sum_{j=1}^{N}B_j$. $q_1,q_2,...,q_N$ are initialized to zeros to let all categories are uniformly distributed at the beginning. In practice, considering that in the early training stage, vcGAN-LP has only seen a subset of the training set and the sense of the real distribution is biased, our model starts training $p_i$s after 2000 batches.

\section{Experiments}
\label{experiments}

We test vcGAN on MNIST~\cite{lecun2010mnist}, Fashion MNIST~\cite{xiao2017fashion}, CIFAR-10~\cite{krizhevsky2009learning}, Cartoon Set and Imbalanced Mixture of CelebA~\cite{liu2015deep} and Cartoon Set. In the experiment on MNIST and Fashion MNIST, we test vcGAN's performance on fidelity of generator distribution and speed of convergence. Moreover, we test its robustness of its hyper-parameter $\delta$. The purpose of experiments on CIFAR-10 and Cartoon Set  is to investigate vcGAN's performance on more complex datasets. In the experiment on imbalanced Mixture of CelebA and Cartoon Set, we want to verify if vcGAN can adapt to imbalanced datasets well. In addition, we also investigate the conditional sampling ability of vcGAN.

In all experiments, the official open source WGAN-GP's~\cite{gulrajani2017improved} implement(DCGAN-like network structure) is the baseline, based on which vcGAN is developed. Each experiment was iterated 200,000 times (1 update for the generator and 5 for the discriminator per iteration), and every 10,000 iterations we sample and save 50,000 samples of the model. The saved samples are used to calculate the widely-used metric called Fr\'echet Inception Distance (FID)~\cite{heusel2017gans} which reflects the difference between the generator distribution and the real data distribution in feature space. The Inception Score (IS)~\cite{salimans2016improved} is also calculated in experiments trained on CIFAR-10. In order to avoid statistical errors, every model of the same configuration was randomly initialized and trained for at least five times to report the mean performance and its standard deviation. To save training time as we have to test many models of different configurations on several datasets for many times, we halved the number of the filters in the DCGAN's generator and discriminator for all the dataset except CIFAR-10. The default $\delta$ was set to 2.0 unless otherwise stated. For more details, please see our codes.

\subsection{Performance of vcGAN}

\subsubsection{MNIST and Fashion MNIST}

The MNIST dataset contains 60,000 gray-scale handwritten digital pictures with a resolution of 28$\times$28. The MNIST data distribution contains 10 evenly distributed patterns, i.e. numbers 0-9. Fashion MNIST has the same data format as MNIST and contains 10 types of evenly distributed images of clothes, pants and shoes. Considering code reuse, we fill the image of the two datasets to a resolution of 32$\times$32 and copy the grayscale channels to form an RGB 3-channel image. 

\begin{figure}[ht]
\begin{center}
\centerline{\includegraphics[width=7.4cm]{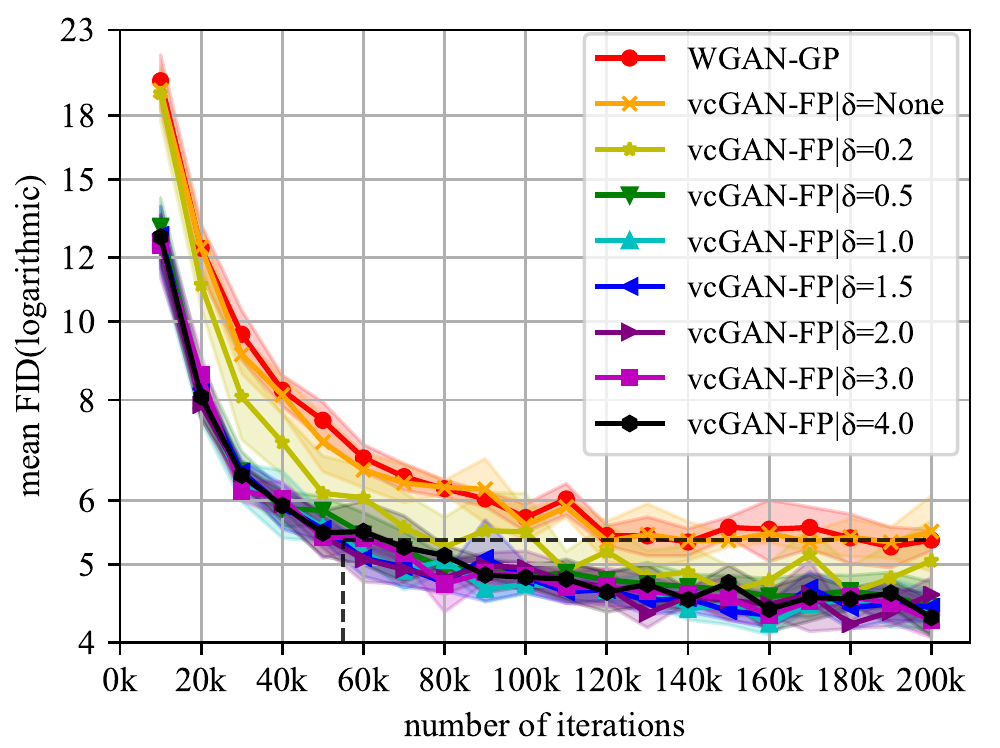}}
\caption{Mean FID (solid line) surrounded by a shaded $\pm\sigma$ area over 5 runs for WGAN-GP and vcGAN-FP with different $\delta$ on MNIST. Lower is better. vcGAN-FP's speed of convergence is about 3.5x that of WGAN-GP on MNIST.}
\label{fid-mnist-pdf}
\end{center}
\vskip -0.1in
\end{figure}

\begin{figure}[ht]
\begin{center}
\centerline{\includegraphics[width=7.4cm]{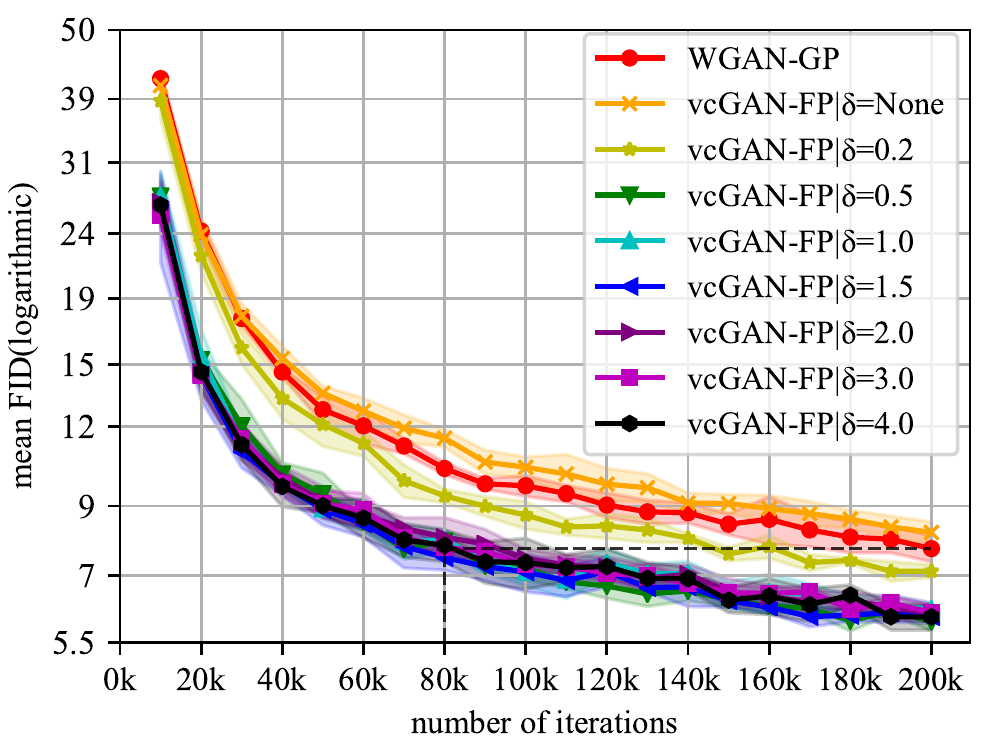}}
\caption{Mean FID (solid line) surrounded by a shaded $\pm1\sigma$ area over 5 runs for WGAN-GP and vcGAN-FP with different $\delta$ on Fashion MNIST. Lower is better. vcGAN-FP's speed of convergence is about 2.5x that of WGAN-GP on Fashion MNIST.}
\label{fid-fashion-pdf}
\end{center}
\vskip -0.3in
\end{figure}

First, we trained a number of vcGAN-FP models with fixed class probability and 10 generative paths, since we already know that real data distributions have 10 patterns of equal proportions. On each dataset, we tried different hyper-parameters $\delta\in[None,0.2,0.5,1,1.5,2,3,4]$, noting that $None$ represents one-hot vectors spliced to Gaussian noise $z'$ without amplification and migration. FID of the generator distribution and training set distribution of each model is shown in Figure~\ref{fid-mnist-pdf} and Figure~\ref{fid-fashion-pdf}.

According to  Figure~\ref{fid-mnist-pdf} and Figure~\ref{fid-fashion-pdf}, when $\delta\geq0.5$, vcGAN-FP learns faster than WGAN-GP since its FID is lower at the same number of iterations. We want to mention that the time per iteration is the same for WGAN-GP and vcGAN-FP. In addition, vcGAN-FP converges to a lower FID value, which means that the vcGAN generator distribution is more faithful to the true distribution. FID of vcGAN-FP is not significantly improved compared to WGAN-GP when the one-hot vector is not amplified, or the degree of amplification is too small ($\delta=0.2$), which indicate that the one-hot vector must be amplified to a sufficient amplitude to enlarge the  euclidean distance between different classes of noise $z_i$ and $z_j$ inputted into $R$. When $\delta=0.5,1,1.5,2,3,4$, the difference of FID curve of vcGAN is not large, indicating that vcGAN is robust to $\delta$. 

Secondly, we trained a number of vcGAN-FP and vcGAN-LP models with a fixed $\delta$ of 3 or 2 on MNIST and Fashion MNIST,respectively, and the number of generative paths $N$ is selected in $[3, 4, 5, 9, 16, 32, 64]$, to simulate the situation when we don't know how many modes are in the training set. Final FID denoted as FID in Table~\ref{fid-mnist} and Table~\ref{fid-fashion} after 200k iterations of each model and the minimum value (Early-stopped FID, ES-FID) calculated every 10k iterations during training are shown in Table~\ref{fid-mnist} and Table~\ref{fid-fashion}.

\begin{table}[t]
\caption{FID of WGAN-GP and vcGAN on MNIST.}
\label{fid-mnist}
\begin{center}
\begin{small}
\begin{sc}
\begin{tabular}{lcccc}
\toprule
Model&$N$&FID &ES-FID\\
\midrule
WGAN-GP	&1	&5.36 	$\pm$0.27&	5.00 	$\pm$0.10\\
vcGAN-FP	&3	&5.01 	$\pm$0.21	&4.78 	$\pm$0.16\\
vcGAN-LP	&3	&4.82 	$\pm$0.29	&4.55 	$\pm$0.11\\
vcGAN-FP	&4	&4.74 	$\pm$0.26	&4.55 	$\pm$0.13\\
vcGAN-LP	&4	&4.94 	$\pm$0.23	&4.48 	$\pm$0.10\\
vcGAN-FP	&5	&4.78 	$\pm$0.17	&4.67 	$\pm$0.10\\
vcGAN-LP	&5	&4.64 	$\pm$0.25	&4.48 	$\pm$0.20\\
vcGAN-FP	&9	&4.64 	$\pm$0.19	&4.36 	$\pm$0.16\\
vcGAN-LP	&9	&4.52 	$\pm$0.21	&4.27 	$\pm$0.18\\
vcGAN-FP	&16	&4.26 	$\pm$0.10	&4.23 	$\pm$0.07\\
vcGAN-FP	&32	&{\bf4.25 	$\pm$0.19}	&{\bf4.09 	$\pm$0.19}\\
vcGAN-FP	&64	&4.42 	$\pm$0.21	&4.17 	$\pm$0.07\\
vcGAN-FP	&10	&4.27 	$\pm$0.21	&4.17 	$\pm$0.09\\
\bottomrule
\end{tabular}
\end{sc}
\end{small}
\end{center}
\vskip -0.1in
\end{table}

\begin{table}[t]
\caption{FID of WGAN-GP and vcGAN on Fashion MNIST.}
\label{fid-fashion}
\begin{center}
\begin{small}
\begin{sc}
\begin{tabular}{lcccc}
\toprule
Model&$N$&FID &ES-FID\\
\midrule
WGAN-GP	&1	&7.72 	$\pm$0.36	&7.72 	$\pm$0.36\\
vcGAN-FP	&3	&6.98 	$\pm$0.11	&6.98 	$\pm$0.11\\
vcGAN-LP	&3	&6.69 	$\pm$0.39	&6.66 	$\pm$0.33\\
vcGAN-FP	&4	&6.96 	$\pm$0.31	&6.90 	$\pm$0.27\\
vcGAN-LP	&4	&6.52 	$\pm$0.19	&6.52 	$\pm$0.19\\
vcGAN-FP	&5	&6.61 	$\pm$0.17	&6.56 	$\pm$0.12\\
vcGAN-LP	&5	&6.40 	$\pm$0.30	&6.40 	$\pm$0.30\\
vcGAN-FP	&9	&6.40 	$\pm$0.34	&6.28 	$\pm$0.32\\
vcGAN-LP	&9	&6.06 	$\pm$0.24	&6.06 	$\pm$0.24\\
vcGAN-FP	&16	&5.95 	$\pm$0.11	&5.92 	$\pm$0.10\\
vcGAN-FP	&32	&5.65 	$\pm$0.07	&5.64 	$\pm$0.06\\
vcGAN-FP	&64	&{\bf5.40 	$\pm$0.19}	&{\bf5.37 	$\pm$0.15}\\
vcGAN-FP	&10	&6.13 	$\pm$0.18	&6.06 	$\pm$0.15\\
\bottomrule
\end{tabular}
\end{sc}
\end{small}
\end{center}
\vskip -0.1in
\end{table}

As shown in Table~\ref{fid-mnist} and Table~\ref{fid-fashion} , vcGAN has lower FIDs on MNIST and Fashion MNIST than WGAN-GP, regardless of whether or not the ADC is allowed to learn the class probability. FID of vcGAN-LP model with the same $N$ is lower than that of vcGAN-FP, which indicates that ADC's learning of categorical probabilities can further improve the quality of vcGAN's generation. We want to mention that vcGAN-LP increased less than 2\% time per iteration compared to WGAN-GP or vcGAN-FP, which is negligible. The generation quality of vcGAN with $N$ less than the number of categories in the training set is inferior to the vcGAN model with $N$ equal to the number of categories in the training set (oracle model), but vcGAN-FP can achieve a lower FID than the orcale model when $N$ is much larger than the number of categories in the real distribution (11st row in Table~\ref{fid-mnist} and 12nd row in Table~\ref{fid-fashion}), without the training of ADC. The reason for this improvement probably is that the model learns to subdivide the classes in the training distribution into more sub-classed, and the distribution of each class is the ensemble of multiple sub-class generative paths.

\subsubsection{CIFAR-10}

CIFAR-10 dataset consists of 60,000 color images of size 32$\times$32 with 10 classes (aircraft, cars, boats, horses, dogs, etc.). We train WGAN-GP and vcGAN-FP with 10 generative paths on CIFAR-10. The IS and FID values for the generator distribution are shown in Table~\ref{cifar-10}.

\begin{table}[t]
\caption{IS and FID of vcGAN-P with different $\delta$ on CIFAR-10.}
\label{cifar-10}
\begin{center}
\begin{small}
\begin{sc}
\begin{tabular}{lccccc}
\toprule
Model&$N$&$\delta$&IS &FID\\
\midrule
WGAN-GP	&1	&	$None$&6.45 	$\pm$0.08	&34.95 $\pm$0.47	\\
vcGAN-FP	&10	&	$None$&6.45 	$\pm$0.03	&34.59 $\pm$0.50	\\
vcGAN-FP	&10	&0.5	&6.55 	$\pm$0.09	&33.70 $\pm$0.53	\\
vcGAN-FP	&10	&1	&6.57 	$\pm$0.07	&33.52 	$\pm$0.53	\\
vcGAN-FP	&10	&2	&6.55 	$\pm$0.05	&33.68 	$\pm$0.47	\\
vcGAN-FP &10 &3 &{\bf6.59 $\pm$0.06}&{\bf 33.43 $\pm$0.61}\\
\bottomrule
\end{tabular}
\end{sc}
\end{small}
\end{center}
\vskip -0.2in
\end{table}

It can be seen from the experimental results that vcGAN-FP outperforms WGAN-GP on CIFAR-10. Although other GAN variants may have higher ISs and lower FIDs than vcGAN, we are not going to compare them, because there are so many variables such as objectives (original GAN loss or Wasserstein loss), network type (DCGAN -like or ResNet-like), number of generators and model size uncontrolled that make the comparison unfair and meaningless. It should be noted that our vcGAN-FP is able to improve the performance of the basic WGAN-GP without increasing the model parameters, changing the training objective function or increasing the calculation. We believe our approach can be plugged into any other GAN variants to boost their performance.

\subsubsection{Cartoon Set}

Cartoon Set is Google’s open source cartoon face avatar dataset (\url{https://google.github.io/cartoonset/download.html}), containing 100,000 PNG format images of size 500$\times$500$\times$4 (including alpha channel). Each cartoon face is composed of several randomly-chosen facial components drawn by the same artist. Every component has discrete variations in color and shape. In this experiment, the alpha channel of the original image is discarded, and pixels between the 123rd to 389th lines and the 131st to 415th columns in the image are clipped and scaled to a size of 64$\times$64$\times$3. We train WGAN-GP and vcGAN-FP models with generative paths of $[2, 4, 10, 16, 32, 64]$. The FID values of the generator distribution are shown in Table~\ref{cartoon}.

\begin{table}[t]
\caption{ FID of  vcGAN-P with different $N$ on Cartoon Set.}
\label{cartoon}
\begin{center}
\begin{small}
\begin{sc}
\begin{tabular}{lcccc}
\toprule
Model&$N$&$\delta$&FID &ES-FID\\
\midrule
WGAN-GP	&1	&$None$	&12.82 	$\pm$0.24	&12.82 	$\pm$0.24	\\
vcGAN-FP	&2	&2	&13.06 	$\pm$0.48	&12.95 $\pm$0.47	\\
vcGAN-FP	&4	&2	&10.89 	$\pm$0.61	&10.82 $\pm$0.57	\\
vcGAN-FP	&10	&2	&11.61 	$\pm$2.15	&11.47 	$\pm$2.10	\\
vcGAN-FP	&16	&2	&11.24 	$\pm$1.86	&11.14 	$\pm$1.65	\\
vcGAN-FP	&32	&2	&{\bf9.57	$\pm$1.11	}&{\bf9.53 	$\pm$1.04}	\\
vcGAN-FP	&64	&2	&10.06 	$\pm$0.58	&10.02 $\pm$0.60	\\
\bottomrule
\end{tabular}
\end{sc}
\end{small}
\end{center}
\vskip -0.2in
\end{table}

Except that the generation quality of vcGAN-FP with 2 generation paths is slightly worse than the baseline, all the vcGAN-FP models outperform WGAN-GP, and the models with $N=32$ and $64$ are significantly beyond the baseline. We owe credit to the ADC module that converts the input analog Gaussian noise into a discrete/digital noise, making the generator of vcGAN more suitable for learning datasets that have discrete features such as Cartoon Set, than the continuous generator in the normal GANs.

\subsubsection{Imbalanced Mixture of CelebA and Cartoon Set}

We mixed CelebA, a dataset of 202,599 celebrity face photos of size 216$\times$178 provided by cropped and aligned, with Cartoon Set unevenly, to simulate imbalanced face images distribution of diverse content and styles on the Internet. In this work, we use the DCGAN's official code (lua version) to further crop and scale the face images of CelebA to a size of 64$\times$64. We produce two versions of the mixed dataset, and the mixture ratios of CelebA and Cartoon Set are 0.7:0.3 ($Ce_7Ca_3$) and 0.8:0.2 ($Ce_8Ca_2$), respectively. WGAN-GP, vcGAN-FP and vcGAN-LP models are trained on these two datasets of imbalanced classes. FID values of the generator distributions are shown in Table~\ref{cartoon_celeba}.

\begin{table}[t]
\caption{FID of WGAN-GP and vcGAN on imbanced face datasets}
\label{cartoon_celeba}
\begin{center}
\begin{small}
\begin{sc}
\begin{tabular}{lccccc}
\toprule
Data&Model&$N$&FID &ES-FID\\
\midrule
$Ce_7Ca_3$	 &WGAN-GP	&1	&21.43 	$\pm$1.15	&20.80 	$\pm$0.82\\
$Ce_7Ca_3$	 &vcGAN-FP	&2	&19.45 	$\pm$0.63	&19.15 	$\pm$0.62\\
$Ce_7Ca_3$	 &vcGAN-LP	&2	&{\bf18.75 	$\pm$0.96	}&{\bf18.41 	$\pm$1.17}\\
$Ce_8Ca_2$	 &WGAN-GP	&1	&19.60 	$\pm$1.29	&19.40 	$\pm$1.34\\
$Ce_8Ca_2$	 &vcGAN-FP	&2	&17.94 	$\pm$0.50	&17.94 	$\pm$0.50\\
$Ce_8Ca_2$	 &vcGAN-LP	&2	&{\bf17.86 	$\pm$2.61}	&{\bf17.09 	$\pm$0.85}\\
\bottomrule
\end{tabular}
\end{sc}
\end{small}
\end{center}
\vskip -0.3in
\end{table}

It can be seen that whether or not the ADC learns the class probability, vcGAN outperforms WGAN-GP in $Ce_7Ca_3$ and $Ce_8Ca_2$. Generators with ADC learning class probability achieves better generation quality.

\subsection{Byproduct --- Conditional Sampling Ability}

\begin{figure}[ht]
\begin{center}
\centerline{\includegraphics[width=7.5cm]{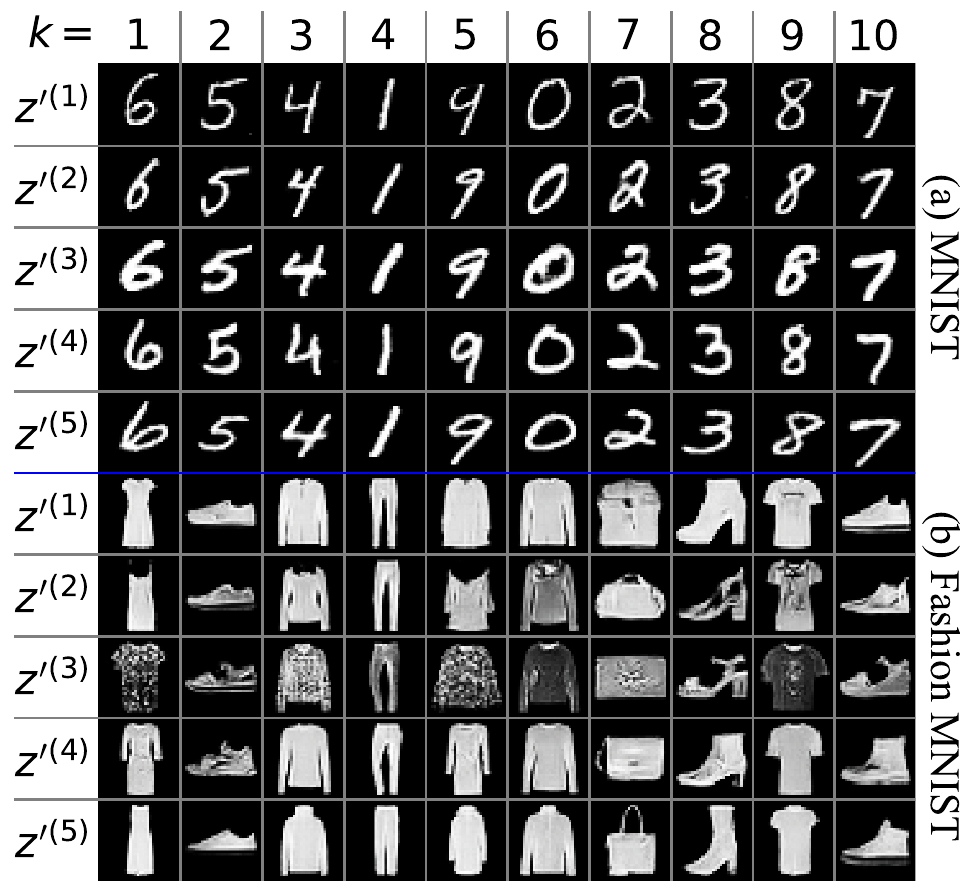}}
\caption{Random subset of vcGAN's conditional samples trained on (a)MNIST and (b)Fashion MNIST. Samples of each column are of the same generative path/category.}
\label{mnist-fashion-conditional}
\end{center}
\vskip -0.3in
\end{figure}

\begin{figure}[ht]
\begin{center}
\centerline{\includegraphics[width=7.5cm]{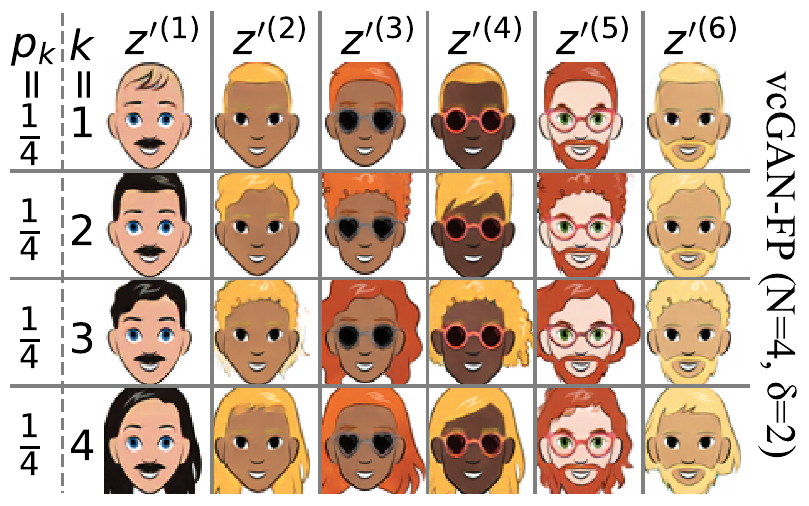}}
\caption{Conditional samples of vcGAN-FP on Cartoon Set. Each generative path yields faces of different discrete amount of hair.}
\label{cartoon-conditional}
\end{center}
\vskip -0.2in
\end{figure}

\begin{figure}[ht]
\begin{center}
\centerline{\includegraphics[width=7cm]{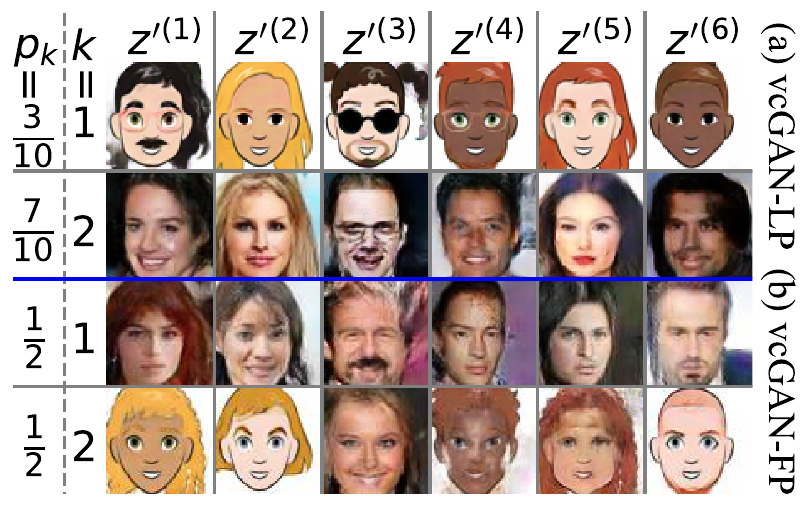}}
\caption{Conditional samples of (a)vcGAN-LP and (b)vcGAN-FP trained on $Ce_7Ca_3$. Training the $p_i$s results in accurate path-wise specialization and better image quality,while the vcGAN-FP's second generative path has to generates both CelebA and Comic faces to create the correct mixture, which causes blur, conflation of modes and higher FID than vcGAN-LP(row 2 and 3 in Table~\ref{cartoon_celeba}).}
\label{ce7ca3-conditional}
\end{center}
\vskip -0.3in
\end{figure}

Training vcGAN on an unlabeled dataset not only yields an unconditional generation model, but also a cGAN that can be sampled by class, which can be realized simply by choosing a specific generative path and inputing the Gaussian noise $z'$. Figures~\ref{mnist-fashion-conditional},~\ref{cartoon-conditional}, and~\ref{ce7ca3-conditional} are subsets of  conditional samples of vcGAN on (MNIST, Fashion MNIST), Cartoon Set, and $Ce_7Ca_3$, respectively. It can be seen that each generative path of vcGAN learns different modes in the real distribution. The discrete noise (virtual label) is responsible for controlling the discrete variations of the generated samples while the analog noise (Gaussian noise $z'$) is responsible for controlling the continuous styles like color, lightness and so on. However, we find vcGANs fail to assign distinct modes to each generative path and ignores the change of $z''$($c$) when $\delta=None$ or less than 0.2, due to insufficient magnitude of the one-hot vector. Results of this case can by found in Appendix.

\begin{figure}[ht]
\begin{center}
\centerline{\includegraphics[width=7.7cm]{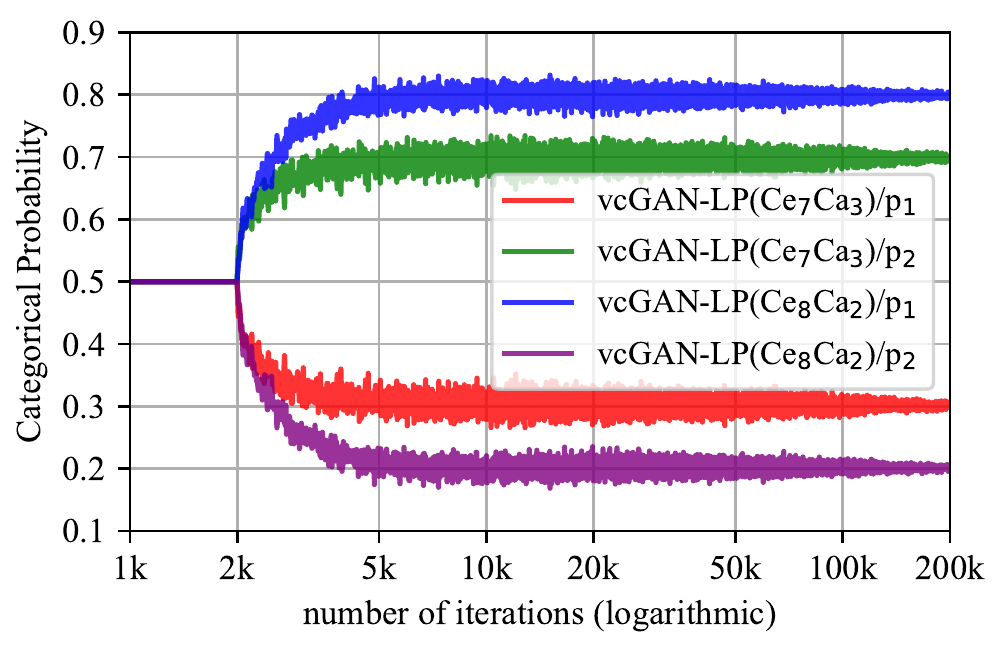}}
\caption{The categorical probabilities of the ADC in vcGAN-LP trained on imbalanced dataset $Ce_7Ca_3$ and $Ce_8Ca_2$. Training of $p_i$s starts from 2k iterations and convergence takes less than 8k iterations. Curves were smoothed by 9-point moving average for better visualization purpose.}
\label{ceca}
\end{center}
\vskip -0.4in
\end{figure}

\section{Discussion}
\label{discussion}

Ordinary GANs can not model the multimodal dataset well, because the mapping of the input Gaussian distribution to the multimodal distribution supported on disconnected sets should have some switch points where the gradient becomes infinite. Such mapping does not belong to the function space of neural network. In order to learn the multimodal distribution, the generator of the ordinary GAN can only approximate the infinite gradient by increasing the gradient of the neural network which consumes the number of iterations and wastes the network's capacity. The cGAN and AC-GAN have input of one-hot category labels injecting a step signal of infinite gradient, which is the patial reason why labels aids. The vcGAN generator has a discontinuous ADC structure that provides a step signal to the decoder network (renderer), avoiding the decoder network wasting the number of iterations and network capacity to approximate the infinite gradient. Therefore, the model distribution's fidelity of cGAN and vcGAN is better than that of normal GAN.

Why can vcGAN's different generative paths learn the different modes automatically with only adversary loss? We guess that when the one-hot vector is amplified, the label-mode aligned mapping is easier than any other unaligned mapping for the decoder network to learn, and is the optimal transportation map~\cite{lei2019geometric} between the latent space and the multimodal image space. Some ensemble GANs (MAD-GAN~\cite{ghosh2018multi} and MGAN~\cite{hoang2017multi}) employ explicit objectives to force different generative paths (generators) to learn separate modes. But the generator may be distracted by the extra objective and exaggerate the inter-mode difference. For example, MGAN has a parameter $\beta$ to balance the  adversary objective and the classification objective and their experiments show that sample quality is very sensitive to $\beta$ ($\beta=0.01$ for CIFAR-10 and 1.0 for STL-10 dataset). Our vcGAN is trained with only the adversary objective, which ensures fidelity to the real data distribution and path-wise specialization is the training \emph{byproduct} rather than purpose. The ADC module which learns the categorical probabilities is also the key to generative paths’specialization. The ADC module is like the gating network in MEGAN, but the gating network is built upon the Gumbel-Softmax which is an approximation and needs annealing to train, while our ADC is based on the Gumbel-Max trick (non-softmax) and no annealing is needed. Moreover, MEGAN's Gating Network selects a sample after all the generators' computation and wastes the unselected $N-1$ samples,  while vcGAN's ADC determines which generative path to draw sample from before the decoder/renderer's computation and is more efficient. Therefore, the propsed vcGAN can learn different modes automatically with only adversary loss.  
 
Some may think vcGAN resembles DeLiGAN which is a lightweight model with multiple generative paths. However, DeLiGAN does not support learning the categorical probabilities and no path-wise specialization was observed. Moreover, vcGAN-FP adds 0 trainable parameter while DeLiGAN adds M$\times$N$\times$2 additional trainable parameters, though still less than other ensemble GANs. It enables vcGAN to equip with 64 generative paths or more for free, which is computationally intractable for existing ensemble GANs.

There are several future works including plugging ADC module into existing ensemble GAN variants or introducing one-hot amplification to cGAN or InfoGAN. One may convert vcGAN to a clustering algorithm and investigate the clustering performance. Experiments can also be conducted to explore if there is any improvement to let the generator learn the $\delta$, A or b, and to see whether the $\delta$ will decay automatically when the training set is not multimodal(only contains one mode).

\section{Conclusion}
\label{conclusion}

This paper proposes a novel GAN model called virtual condition GAN, which has multiple generative paths with a shared decoder network and a learnable ADC yielding virtual labels. The vcGAN has both merits of ensemble GANs and conditional GANs and only need adversary loss and unlabeled dataset to train with. Many experiments on several balanced/imbalanced image datasets indicate that vcGAN can generate better samples, achieve lower FID, have faster convergence and robustness to hyperparameters. What's more, it supports class-conditional sampling.

\bibliography{example_paper}
\bibliographystyle{icml2019}
%
%
%
%
%

\end{document}